\documentclass[11pt,a4paper]{article}
\usepackage[hyperref]{acl2020}
\usepackage{times}
\usepackage{latexsym}
\usepackage{amsmath}
\usepackage{bbm}
\usepackage{tabularx}
\usepackage[ruled,vlined]{algorithm2e}
\DeclareMathOperator*{\argmax}{argmax}

\usepackage{graphicx}
\usepackage{amsfonts}
\usepackage{hyperref}
\SetKwProg{Fn}{Function}{}{}
\usepackage{microtype}

\aclfinalcopy 

\title{Calibrating Structured Output Predictors for Natural Language Processing}

\author{Abhyuday Jagannatha$^{1}$, Hong Yu$^{1,2}$ \\
$^1$College of Information and Computer Sciences,University of Massachusetts Amherst 
\\$^2$Department of Computer Science,University of Massachusetts Lowell \\
\texttt{\{abhyuday, hongyu\}@cs.umass.edu}
}

\date{}

\begin{document}
\maketitle
\begin{abstract}
We address the problem of calibrating prediction confidence for output entities of interest in natural language processing (NLP) applications. It is important that NLP applications such as named entity recognition and question answering produce calibrated confidence scores for their predictions, especially if the applications are to be deployed in a safety-critical domain such as healthcare. However, the output space of such structured prediction models is often too large to adapt binary or multi-class calibration methods directly. In this study, we propose a general calibration scheme for output entities of interest in neural network based structured prediction models. Our proposed method can be used with any binary class calibration scheme and a neural network model. Additionally, we show that our calibration method can also be used as an uncertainty-aware, entity-specific decoding step to improve the performance of the underlying model at no additional training cost or data requirements. We show that our method outperforms current calibration techniques for named-entity-recognition, part-of-speech and question answering. We also improve our model's performance from our decoding step across several tasks and benchmark datasets. Our method improves the calibration and model performance on out-of-domain test scenarios as well. 
\end{abstract}

\section{Introduction}

Several modern machine-learning based Natural Language Processing (NLP) systems can provide a confidence score with their output predictions. This score can be used as a measure of predictor confidence. A well-calibrated confidence score is a probability measure that is closely correlated with the likelihood of model output's correctness. As a result, NLP systems with calibrated confidence can predict when their predictions are likely to be incorrect and therefore, should not be trusted. This property is necessary for the responsible deployment of NLP 
systems in safety-critical domains such as healthcare and finance. Calibration of predictors is a well-studied problem in machine learning \citep{guo2017calibration,platt1999probabilistic}; however, widely used methods in this domain are often defined as binary or multi-class problems\citep{naeini2015obtaining,nguyen2015posterior}. The structured output schemes of NLP tasks such as information extraction (IE) \citep{sang2003introduction} and extractive question answering \citep{rajpurkar2018know} have an output space that is often too large for standard multi-class calibration schemes.


Formally, we study NLP models that provide conditional probabilities $p_\theta(y|x)$ for a structured output $y$ given input $x$. The output can be a label sequence in case of part-of-speech (POS) or named entity recognition (NER) tasks, or a span prediction in case of extractive question answering (QA) tasks, or a relation prediction in case of relation extraction task. $p_\theta(y|x)$ can be used as a score of the model's confidence in its prediction. However, $p_\theta(y|x)$  is often a poor estimate of model confidence for the output $y$. The output space of the model in sequence-labelling tasks is often large, and therefore $p_\theta(y|x)$ for any output instance $y$ will be small. For instance, in a sequence labelling task with $C$ number of classes and a sequence length of $L$, the possible events in output space will be of the order of $C^L$. Additionally, recent efforts \citep{guo2017calibration,nguyen2015posterior,dong2018confidence,kumar2019calibration} at calibrating machine learning models have shown that they are poorly calibrated. Empirical results from \citet{guo2017calibration} show that techniques used in deep neural networks such as dropout and their large architecture size can negatively affect the calibration of their outputs in binary and multi-class classification tasks.  

Parallelly, large neural network architectures based on contextual embeddings \citep{devlin2018bert,peters2018deep} have shown state-of-the-art performance across several NLP tasks \citep{andrew2007scalable,wang2019superglue} . They are being rapidly adopted for information extraction and other NLP tasks in safety-critical applications \citep{zhu2018clinical,sarabadani2019detection,fei2019BERT,lee2019biobert}. Studying the miss-calibration in such models and efficiently calibrating them is imperative for their safe deployment in the real world.

In this study, we demonstrate that neural network models show high calibration errors for NLP tasks such as POS, NER and QA. We extend the work by \citet{kuleshov2015calibrated} to define well-calibrated forecasters for output entities of interest in structured prediction of NLP tasks. We provide a novel calibration method that applies to a wide variety of NLP tasks and can be used to produce model confidences for specific output entities instead of the complete label sequence prediction. We provide a general scheme for designing manageable and relevant output spaces for such problems.  We show that our methods lead to improved calibration performance on a variety of benchmark NLP datasets. Our method also leads to improved out-of-domain calibration performance as compared to the baseline, suggesting that our calibration methods can generalize well. 

Lastly, we propose a procedure to use our calibrated confidence scores to re-score the predictions in our defined output event space. This procedure can be interpreted as a scheme to combine model uncertainty scores and entity-specific features with decoding methods like Viterbi. We show that this re-scoring leads to consistent improvement in model performance across several tasks at no additional training or data requirements. 

\section{Calibration framework for Structured Prediction NLP models}
\subsection{Background}
Structured Prediction refers to the task of predicting a structured output $y=[y_1,y_2,...y_L]$ for an input $x$. In NLP, a wide array of tasks including parsing, information extraction, and extractive question answering fall within this category. Recent approaches towards solving such tasks are commonly based on neural networks that are trained by minimizing the following objective : 
\begin{equation}
    \mathcal{L}(\theta|\mathcal{D})=-\sum_{i=0}^{|\mathcal{D}|}log(p_\theta(y^{(i)}|x^{(i)})) + R(\theta) 
\end{equation}
where $\theta$ is the parameter vector of the neural network and $R$ is the regularization penalty and $\mathcal{D}$ is the dataset $\{(y^{(i)},x^{(i)})\}_{i=0}^{|\mathcal{D}|}$. The trained model $p_\theta$ can then be used to produce the output $\hat{y}=\argmax_{y \in \mathcal{Y}} p_\theta(y|x)$. Here, the corresponding model probability $p_\theta(\hat{y}|x)$ is the uncalibrated confidence score.

In binary class classification, the output space $\mathcal{Y}$ is $[0,1]$. The confidence score for such classifiers can then be calibrated by training a forecaster $F_y:[0,1] \to [0,1]$ which takes in the model confidence $F_y(P_\theta(y|x))$ to produce a recalibrated score \citep{platt1999probabilistic}. A widely used method for binary class calibration is Platt scaling where $F_y$ is a logistic regression model. Similar methods have also been defined for multi-class classification \citep{guo2017calibration}. However, extending this to structured prediction in NLP settings is non-trivial since the output space $|\mathcal{Y}|$ is often too large for us to calibrate the output probabilities of all events. 
\subsection{Related Work}
Calibration methods for binary/multi class classification has been widely studied in related literature \citep{brocker2009reliability,guo2017calibration}. Recent efforts at confidence modeling for NLP has focused on several tasks like co-reference, \citep{nguyen2015posterior}, semantic parsing \citep{dong2018confidence} and neural machine translation \citep{kumar2019calibration}.


\subsection{Calibration in Structured Prediction}

In this section, we define the calibration framework by \citet{kuleshov2015calibrated} in the context of structured prediction problems in NLP.  The model $p_\theta$ denotes the neural network that produces an conditional probability $p_\theta(y|x)$ given an $(x,y)$ tuple. In a multi/binary class setting, a function $F_y$ is used to map the output $p_\theta(y|x)$ to a calibrated confidence score for all $y \in \mathcal{Y}$. In a structured prediction setting, since the cardinality of $\mathcal{Y}$ is usually large, we instead focus on the event of interest set $\mathcal{I}(x)$. $\mathcal{I}(x)$ contains events of interest $E$ that are defined using the output events relevant to the deployment requirements of a model. The event $E$ is a subset of $\mathcal{Y}$ . There can be several different schemes to define $\mathcal{I}(x)$. In later sections, we discuss related work on calibration that can be understood as applications of different $\mathcal{I}(x)$ schemes. In this work, we define a general framework for constructing $\mathcal{I}(x)$ for NLP tasks which allows us to maximize calibration performance on output entities of interest.

We define $F_y(E,x,p_\theta)$ to be a function, that takes the event $\textit{E}$, the input feature $x$ and $p_\theta$ to produce a confidence score between $[0,1]$. We refer to this calibration function as the forecaster and use $F_y(E,x)$ as a shorthand since it is implicit that $F_y$ depends on outputs of $p_\theta$. We would like to find the forecaster that minimizes the discrepancy between $F_y(\textit{E},x)$ and $\mathbb{P}(y\in \textit{E}|x)$ for $(x,y)$ sampled from $\mathbb{P}(x,y)$ and $\textit{E}$ uniformly sampled from $\mathcal{I}(\textit{x})$.

A commonly used methodology for constructing a forecaster for $p_\theta$ is to train it on a held-out dataset $D_{dev}$. A forecaster for a binary classifier is perfectly calibrated if
\begin{equation}
    \mathbb{P}(y=1|F_y(x)=p) = p .
\end{equation}

It is trained on samples from $\{(x,\mathbb{I}(y=1):(x,y) \in D_{dev}\}$. For our forecaster based on $\mathcal{I}(x)$, perfect calibration would imply that
\begin{equation}
    \mathbb{P}(y\in E|F_y(x,\textit{E})=p) = p .
\end{equation}

The training data samples for our forecaster are $\{(x,\mathbb{I}(y \in \textit{E}) : \textit{E} \in \mathcal{I}(x), (x,y) \in D_{dev} \}$. 

\subsection{Construction of Event of Interest set $\mathcal{I}(x)$}
\label{sec:eventofinterest}
The main contributions of this paper stem from our proposed schemes for constructing the aformentioned $\mathcal{I}(x)$ sets for NLP applications.

\paragraph{Entities of Interest}:
In the interest of brevity, let us define ``Entities of interest'' $\phi(x)$ as the set of all entity predictions that can be queried from $p_\theta$ for a sample $x$. For instance, in the case of answer span prediction for QA, the $\phi(x)$ may contain the MAP prediction of the best answer span (answer start and end indexes). In a parsing or sequence labeling task, $\phi(x)$ may contain the top-k label sequences obtained from viterbi decoding. In a relation or named-entity extraction task, $\phi(x)$ contains the relation or named entity span predictions respectively. Each entity $s$ in $\phi(x)$ corresponds to a event set $\textit{E}$ that is defined by all outputs in $\mathcal{Y}$ that contain the entity $s$.  $\mathcal{I}(x)$ contains set $\textit{E}$ for all entities in $\phi(x)$. 

\paragraph{Positive Entities and Events}:
We are interested in providing a calibrated probability for $y\in\textit{E}$ corresponding to an $s$ for all $s$ in $\phi(x)$. Here $y$ is the correct label sequence for the input $x$. If $y$ lies in the set $\textit{E}$ for an entity $s$, we refer to $s$ as a positive entity and the event as a positive event. In the example of named entity recognition, $s$ may refer to a predicted entity span, $\textit{E}$ refers to all possible sequences in $\mathcal{Y}$ that contain the predicted span. The corresponding event is positive if the correct label sequence $y$ contains the span prediction $s$.

\paragraph{Schemes for construction of $\mathcal{I}(x)$}:
While constructing the set $\phi(x)$ we should ensure that it is limited to a relatively small number of output entities, while still covering as many positive events in $\mathcal{I}(x)$ as possible. To explain this consideration, let us take the example of a parsing task such as syntax or semantic parsing. Two possible schemes for defining $\mathcal{I}(x)$ are :
\begin{enumerate}
    \item Scheme 1: $\phi(x)$ contains the MAP label sequence prediction. $\mathcal{I}(x)$ contains the event corresponding to whether the label sequence $y'=\argmax_y p_\theta(y|x)$ is correct.
    \item Scheme 2: $\phi(x)$ contains all possible label sequences. $\mathcal{I}(x)$ contains a event  corresponding to whether the label sequence $y'$ is correct, for all $y' \in \mathcal{Y}$
\end{enumerate}

\begin{table*}[ht]
\centering
\begin{tabular}{p{0.25\linewidth}p{0.13\linewidth}p{0.13\linewidth}p{0.13\linewidth}}
\textbf{Calibration} & \textbf{\textsc{bert}} & \textbf{\textsc{bert}+CRF} & \textbf{Distil\textsc{bert}} \\ \hline \hline
Platt & 15.90$\pm$.03  & 15.56$\pm$.23 & 12.30$\pm$.13 \\
Calibrated Mean & 2.55$\pm$.34  & \textbf{2.31}$\pm$\textbf{.35} & \textbf{2.02}$\pm$\textbf{.16} \\
 +Var & \textbf{2.11}$\pm$\textbf{.32} &2.55$\pm$.32 &2.73$\pm$.40 \\
 \hline
Platt+top2 & 11.4$\pm$.07  & 14.21$\pm$.16 & 11.03$\pm$.31\\
Calibrated Mean+top2 & 2.94$\pm$ .29  & 4.82$\pm$.15 & 3.61$\pm$.17 \\
+Var+top2 & 2.17$\pm$.35  & 4.26$\pm$.10 & 2.43$\pm$.16 \\
+Rank+top2 & 2.43$\pm$.30 & 2.43$\pm$.45 & 2.21$\pm$.09\\
+Rank+Var+top2 & \textbf{1.81}$\pm$\textbf{.12} & \textbf{2.29}$\pm$\textbf{.27} & \textbf{1.97}$\pm$\textbf{.14}\\
\hline
Platt+top3 & 17.46$\pm$.13  & 18.11$\pm$.16 & 12.84$\pm$.37\\
+Rank+Var+top3 & 3.18$\pm$.12 & 3.71$\pm$.25 & 2.05$\pm$.06 \\
\hline
\end{tabular}
\caption{\label{table:PTB-ECE} ECE percentages on Penn Treebank for different models and calibration methods. The results are for top-$1$ MAP predictions on the test data. ECE standard deviation is estimated by repeating the experiments for 5 repetitions. ECE for uncalibrated \textsc{bert}, \textsc{bert}+CRF model and Distil\textsc{bert} is 35.11\%, 33.72\% and 28.06\% respectively. \textit{heuristic-$k$} is 2 for all +Rank+Var+topk forecasters. Full feature model +Rank+Var+topk, $k=3$ is also provided for completeness.}
\end{table*}

Calibration of model confidence by \citet{dong2018confidence} can be viewed as Scheme 1, where the entity of interest is the MAP label sequence prediction. Whereas, using Platt Scaling in a one-vs-all setting for multi-class classification \citep{guo2017calibration} can be seen as an implementation of Scheme 2 where the entity of interest is the presence of class label. As discussed in previous sections, Scheme 2 is too computationally expensive for our purposes due to large value of $|\mathcal{Y}|$ . Scheme 1 is computationally cheaper, but it has lower coverage of positive events. For instance, a sequence labelling model with a 60\% accuracy at sentence level means that only 60 \% of positive events are covered by the set corresponding to $\argmax_y p_\theta(y|x)$ predictions. In other words, only 60 \% of the correct outputs of model $p_\theta$ will be used for constructing the forecaster. This can limit the positive events in $\mathcal{I}(x)$. Including the top-$k$ predictions in $\phi(x)$ may increase the coverage of positive events and therefore increase the positive training data for the forecaster. The optimum choice of $k$ involves a trade-off. A larger value of $k$ implies broader coverage of positive events and more positive training data for the forecaster training. However, it may also lead to an unbalanced training dataset that is skewed in favour of negative training examples.

Task specific details about $\phi(x)$ are provided in the later sections. For the purposes of this paper, top-$k$ refers to the top $k$ MAP sequence predictions, also referred to as  $argmax(k)$.

\subsection{Forecaster Construction}
\label{sec:forecaster}
Here we provide a summary of the steps involved in Forecaster construction. Remaining details are in the Appendix. We train the neural network model $p_\theta$ on the training data split for a task and use the validation data for monitoring the loss and early stopping. After the training is complete, this validation data is re-purposed to create the forecaster training data. We use an MC-Dropout\citep{gal2016dropout} average of (n=10) samples to get a low variance estimate of logit outputs from the neural networks. This average is fed into the decoding step of the model $p_\theta$ to obtain top-$k$ label sequence predictions. We then collect the relevant entities in $\phi(x)$, along with the $\mathbb{I}(y \in \textit{E)}$ labels to form the training data for the forecaster. We use gradient boosted decision trees \citep{friedman2001greedy} as our region-based \citep{dong2018confidence,kuleshov2015calibrated} forecaster model.
\paragraph{Choice of the hyperparameter $k$:}
We limit our choice of k to $\{2,3\}$. We train our forecasters on training data constructed through top-$2$ and top-$3$ extraction each. These two models are then evaluated on top-$1$ extraction training data, and the best value of $k$ is used for evaluation on test. This heuristic for $k$ selection is based on the fact that the top-$1$ training data for a \textit{good} predictor $p_\theta$, is a positive-event rich dataset. Therefore, this dataset can be used to reject a larger $k$ if it leads to reduced performance on positive events. We refer to the value of $k$ obtained from this heuristic as  as \textit{heuristic-$k$}.

\begin{table*}[ht]
\centering
\begin{tabular}{p{0.25\linewidth}p{0.13\linewidth}p{0.13\linewidth}p{0.13\linewidth}}
\textbf{Calibration} & \textbf{\textsc{bert}} & \textbf{\textsc{bert}+CRF} & \textbf{Distil\textsc{bert}} \\ \hline \hline
Baseline & 60.30$\pm$.12  & 62.31$\pm$.11 & 60.17$\pm$.08\\
+Rank+Var+top2 & 60.30$\pm$.23  & 62.31$\pm$.09 & 60.13$\pm$.11\\
+Rank+Var+top3 & 59.84$\pm$.16  & 61.06$\pm$.14 & 58.95$\pm$.08\\\hline
\end{tabular}
\caption{\label{table:PTB-Acc} Micro-avg f-score for POS datasets using the baseline and our best proposed calibration method. The confidence score from the calibration method is used to re-rank the events $\textit{E} \in \mathcal{I}(s)$ and the top selection is chosen. Standard deviation is estimated by repeating the experiments for 5 repetitions. Baseline refers to MC-dropout averaged (sample-size=10) output from the model $p_\theta$. \textit{heuristic-$k$} is 2 for +Rank+Var+topk forecasters. }
\end{table*}

\subsection{Feature Construction for Calibration}
We use three categories of features as inputs to our forecaster.

\textbf{Model and Model Uncertainty based features} contain the mean probability obtained by averaging over the marginal probability of the ``entity of interest'' obtained from 10 MC-dropout samples of $p_\theta$. Average of marginal probabilities acts as a reduced variance estimate of un-calibrated model confidence. Our experiments use the pre-trained contextual word embedding architectures as the backbone networks. We obtain MC-Dropout samples by enabling dropout sampling for all dropout layers of the networks. We also provide $10^{th}$  and $90^{th}$ percentile values from the MC-Dropout samples, to provide model uncertainty information to the forecaster. Since our forecaster training data contains entity predictions from top-$k$ MAP predictions, we also include the rank $k$ as a feature. We refer to these two features as ``Var'' and ``Rank'' in our models.

\textbf{Entity of interest based features} contain the length of the entity span if the output task is named entity. We only use this feature in the NER experiments and refer to it as ``ln''.

\textbf{Data Uncertainty based features:} \citet{dong2018confidence} propose the use of language modelling (LM) and OOV-word-based features as a proxy for data uncertainty estimation. The use of word-pieces and large pre-training corpora in contextual word embedding models like \textsc{bert} may affect the efficacy of LM based features. Nevertheless, we use LM perplexity (referred to as ``lm'') in the QA task to investigate its effectiveness as an indicator of the distributional shift in data. Essentially, our analysis focuses on LM perplexity as a proxy for \textit{distributional} uncertainty \citep{malinin2018predictive} in our out-of-domain experiments. The use of word-pieces in models like \textsc{bert} reduces the negative effect of OOV words on model prediction. Therefore, we do not include OOV features in our experiments. 

\section{Experiments and Results}
We use \textsc{bert}-base \citep{devlin2018bert} and distil\textsc{bert} \citep{sanh2019distilbert} network architecture for our experiments. Validation split for each dataset was used for early stopping \textsc{bert} fine-tuning and as training data for forecaster training.  POS and NER experiments are evaluated on Penn Treebank and CoNLL 2003 \citep{sang2003introduction}, MADE 1.0 \citep{jagannatha2019overview} respectively. QA experiments are evaluated on SQuAD1.1  \citep{rajpurkar2018know} and EMRQA \citep{pampari2018emrqa} corpus. We also investigate the performance of our forecasters on an out-of-domain QA corpus constructed by applying EMRQA QA data generation scheme \citep{pampari2018emrqa} on the MADE 1.0 named entity and relations corpus. Details for these datasets are provided in their relevant sections. 

We use the expected calibration error (ECE) metric defined by \citet{naeini2015obtaining} with $N=20$ bins \citep{guo2017calibration} to evaluate the calibration of our models. ECE is defined as an estimate of the expected difference between the model confidence and accuracy. ECE has been used in several related works \citep{guo2017calibration,maddox2019simple,kumar2018trainable,vaicenavicius2019evaluating} to estimate model calibration. We use Platt scaling as the baseline calibration model. It uses the length-normalized probability averaged across $10$ MC-Dropout samples as the input. The lower variance and length invariance of this input feature make Platt Scaling a strong baseline. We also use a ``Calibrated Mean'' baseline using Gradient Boosted Decision Trees as our estimator with the same input feature as Platt.

\subsection{Calibration for Part-of-Speech Tagging}
Part-of-speech (POS) is a sequence labelling task where the input is a text sentence, and the output is a sequence of syntactic tags. We evaluate our method on the Penn Treebank dataset \cite{marcus1994penn}. 
We can define either the token prediction or the complete sequence prediction as the entity of interest. Since using a token level entity of interest effectively reduces the calibration problem to that of calibrating a multi-class classifier, we instead study the case where the predicted label sequence of the entire sentence forms the entity of interest set.
The event of interest set is defined by the events ${y=\text{MAP}_k(x)}$ which denote whether each top-$k$ sentence level MAP prediction is correct. 
We use three choice of $p_\theta$ models, namely \textsc{bert}, \textsc{bert}-CRF and distil\textsc{bert}.
We use model uncertainty and rank based features for our POS experiments. 

\begin{table}
\centering
\begin{tabular}{lll}
\textbf{Calibration} & \textbf{CoNLL} & \textbf{MADE 1.0} \\ 
 & (\textsc{bert})  & (bio\textsc{bert}) \\ \hline \hline
Platt & \textbf{2.00}$\pm$\textbf{.12}  & 4.00$\pm$.07 \\
Calibrated Mean & 2.29$\pm$.33  & 3.07$\pm$.18 \\
 +Var & 2.43$\pm$.36 &3.05$\pm$.17 \\
 +Var+ln & 2.24$\pm$.14  & \textbf{2.92}$\pm$\textbf{.24}\\\hline
 
Platt+top3 & 16.64$\pm$.48  & 2.14$\pm$.18 \\
Calibrated Mean+top3 & 17.06$\pm$.50  & 2.22$\pm$.31 \\
 +Var+top3 & 17.10$\pm$.24 &2.17$\pm$.39 \\
 +Rank+Var+top3 & 2.01$\pm$.33 & 2.34$\pm$.15\\ 
 +Rank+Var+ln+top3 & \textbf{1.91}$\pm$\textbf{.29}  & \textbf{2.12}$\pm$\textbf{.24}\\ \hline
\end{tabular}
\caption{\label{table:NER-ECE} ECE percentages for the two named entity datasets and calibration methods. The results are for all predicted named entity spans in top-1 MAP predictions on the test data. ECE standard deviation is estimated by repeating the experiments for $5$ repetitions. ECE for uncalibrated span marginals from \textsc{bert} model is 3.68\% and 5.59\% for CoNLL and MADE 1.0 datasets. \textit{heuristic-$k$} is 3 for all +Rank+Var+top3 forecasters.}
\end{table}

Table \ref{table:PTB-ECE} shows the ECE values for our baseline, proposed and ablated models. The  value of \textit{heuristic-$k$} is $2$ for all +Rank+Var+topk forecasters across all PTB models. ``top$k$'' in Table \ref{table:PTB-ECE} refers to forecasters trained with additional top-$k$ predictions. Our methods outperform both baselines by a large margin. Both ``Rank'' and ``Var'' features help in improving model calibration. Inclusion of top-$2$ prediction sequences also improve the calibration performance significantly. Table \ref{table:PTB-ECE} also shows the performance of our full feature model ``+Rank+Var+top$k$'' for the sub-optimal value of $k=3$. It has lower performance than $k=2$ across all models. Therefore for the subsequent experimental sections, we only report top$k$ calibration performance using the \textit{heuristic-$k$} value only.

We use the confidence predictions of our full-feature model +Rank+Var+top$k$ to re-rank the top-$k$ predictions in the test set. Table \ref{table:PTB-Acc} shows the sentence-level (entity of interest) accuracy for our re-ranked top prediction and the original model prediction. 

\begin{table}
\centering
\begin{tabular}{lll}
\textbf{Calibration} & \textbf{CoNLL} & \textbf{MADE 1.0} \\
 & (\textsc{bert})  & (bio\textsc{bert}) \\ \hline \hline

Baseline & 89.45$\pm$.08  & 84.01$\pm$.11 \\
+Rank+Var+top3 & 89.73$\pm$.12  & 84.33$\pm$.07 \\
+Rank+Var+ln+top3 & \textbf{89.78}$\pm$\textbf{.10}  & \textbf{84.34}$\pm$\textbf{.10}\\\hline
 
\end{tabular}
\caption{\label{table:NER-Acc} Micro-avg f-score for NER datasets and our best proposed calibration method. The confidence score from the calibration method is used to re-rank the events $\textit{E} \in \mathcal{I}(s)$ and a confidence value of 0.5 is used as a cutoff. Standard deviation is estimated by repeating the experiments for $5$ repetitions. Baseline refers to MC-dropout averaged (sample-size=10) output of model $p_{\theta}$. \textit{heuristic-$k$} is 3 for all +Rank+Var+top3 forecasters.}
\end{table}
\subsection{Calibration for Named Entities}
For Named Entity (NE) Recognition experiments, we use two NE annotated datasets, namely CoNLL 2003 and MADE 1.0. CoNLL 2003 consists of documents from the Reuters corpus annotated with named entities such as Person, Location etc. MADE 1.0  dataset is composed of electronic health records annotated with clinical named entities such as Medication, Indication and Adverse effects. 

The entity of interest for NER is the named entity span prediction. We define $\phi(x)$ as predicted entity spans in $\text{argmax}(k)$ label sequences predictions for $x$. We use \textsc{bert}-base with token-level softmax output and marginal likelihood based training. 
The model uncertainty estimates for ``Var'' feature are computed by estimating the variance of length normalized MC-dropout samples of span marginals. Due to the similar trends in behavior of \textsc{bert} and \textsc{bert}+CRF model in POS experiments, we only use \textsc{bert} model for NER. However, the span marginal computation can be easily extended to linear-chain CRF models. We also use the length of the predicted named entity as the feature ``ln'' in this experiment. Complete details about forecaster and baselines are in the Appendix. Value of  \textit{heuristic-$k$}  is 3 for all +Rank+Var+topk forecasters. We show ablation and baseline results for $k=3$ only. However, no other forecasters for any $k\in \{2,3\}$ outperform our best forecasters in Table \ref{table:NER-ECE}. 


\begin{table*}[ht]
\centering
\begin{tabular}{p{0.25\linewidth}p{0.13\linewidth}p{0.13\linewidth}p{0.13\linewidth}p{0.13\linewidth}}

\textbf{Calibration} & \textbf{SQuAD1.1} & \textbf{EMRQA} & \textbf{MADE 1.0} &\textbf{MADE 1.0(OOD)}\\
 & (\textsc{bert})  & (bio\textsc{bert}) & (bio\textsc{bert}) & (bio\textsc{bert}) \\ \hline \hline
Platt & 3.69$\pm$.16 &5.07$\pm$.37&3.64$\pm$.17&15.20$\pm$.16\\
Calibrated Mean & 2.95$\pm$.26 &2.28$\pm$.18&2.50$\pm$.31&13.26$\pm$.94\\
+Var & \textbf{2.92}$\pm$\textbf{.28} &2.74$\pm$.15&2.71$\pm$.32&\textbf{12.41}$\pm$\textbf{.95}\\
\hline
Platt+top3 & 7.71$\pm$.28&5.42$\pm$.25&11.87$\pm$.19&16.36$\pm$.26\\
Calibrated Mean+top3 & 3.52$\pm$.35 &2.11$\pm$.19&9.21$\pm$.25&12.11$\pm$.24\\
+Var+top3 & 3.56$\pm$.29 &2.20$\pm$.20&9.26$\pm$.27&\textbf{11.67}$\pm$\textbf{.27}\\
+Var+lm+top3 & 3.54$\pm$.21 &2.12$\pm$.19&6.07$\pm$.26&12.42$\pm$.32\\
+Rank+Var+top3 & \textbf{2.47}$\pm$\textbf{.18} &\textbf{1.98}$\pm$\textbf{.10}&1.77$\pm$.23&12.69$\pm$.20\\
+Rank+Var+lm+top3 & 2.79$\pm$.32 &2.24$\pm$.29&\textbf{1.66}$\pm$\textbf{.27}&12.60$\pm$.28\\
\hline
\end{tabular}
\caption{\label{table:QA-ECE}ECE percentages for QA tasks SQuAD1.1, EMRQA and MADE 1.0. MADE 1.0(OOD) refers to the out-of-domain evaluation of a QA model that is trained and calibrated on EMRQA training and validation splits. The results are for top-1 MAP predictions on the test data. ECE standard deviation is estimated by repeating the experiments for 5 repetitions. \textsc{bert} model's uncalibrated ECE for SQuAD1.1, EMRQA, MADE 1.0 and MADE 1.0(OOD) are 6.24\%  6.10\%, 20.10\% and 18.70\% respectively. \textit{heuristic-$k$} is 3 for all +Rank+Var+top$k$ forecasters.}
\end{table*}

We use the confidence predictions of our ``+Rank+Var+top3'' models to re-score the confidence predictions for all spans predicted in top-$3$ MAP predictions for samples in the test set. A threshold of 0.5 was used to remove span predictions with low confidence scores. Table \ref{table:NER-Acc} shows the Named Entity level (entity of interest) Micro-F score for our re-ranked top prediction and the original model prediction. We see that re-ranked predictions from our models consistently improve the model f-score.

\begin{table*}[ht]
\centering
\begin{tabular}{p{0.25\linewidth}p{0.13\linewidth}p{0.13\linewidth}p{0.13\linewidth}p{0.13\linewidth}}

\textbf{Calibration} & \textbf{SQuAD1.1} & \textbf{EMRQA} & \textbf{MADE 1.0} & \textbf{MADE 1.0(OOD)}\\ & (\textsc{bert})  & (bio\textsc{bert}) & (bio\textsc{bert}) & (bio\textsc{bert}) \\ \hline \hline
Baseline & 79.79$\pm$.08 &70.97$\pm$.14&66.21$\pm$.18&31.62$\pm$.12\\

+Rank+Var+top3 & \textbf{80.04}$\pm$\textbf{.11} &71.34$\pm$.22&\textbf{66.33}$\pm$\textbf{.12}&31.99$\pm$.11\\

+Rank+Var+lm+top3 & \textbf{80.03}$\pm$\textbf{.15}&\textbf{71.37}$\pm$\textbf{.26}& \textbf{66.33}$\pm$\textbf{.15}& \textbf{32.02}$\pm$\textbf{.09}\\
\hline
\end{tabular}
\caption{\label{table:QA-Acc} Table shows change in Exact Match Accuracy for QA datasets and our best proposed calibration method. The confidence score  from the calibration method is used to re-rank the events $\textit{E} \in \mathcal{I}(s)$. Standard deviation is estimated by repeating the experiments for 5 repetitions. Baseline refers to MC-dropout averaged (sample-size=10) output of model $p_\theta$. \textit{heuristic-$k$} is 3 for all +Rank+Var+top$k$ forecasters.}
\end{table*}

\subsection{Calibration for QA Models}
We use three datasets for evaluation of our calibration methods on the QA task. Our QA tasks are modeled as extractive QA methods with a single span answer predictions. We use three datasets to construct experiments for QA calibration. SQuAD1.1 and EMRQA \citep{pampari2018emrqa} are open-domain and clinical-domain QA datasets, respectively. We process the EMRQA dataset by restricting the passage length and removing unanswerable questions. We also design an out-of-domain evaluation of calibration using clinical QA datasets. We follow the guidelines from \citet{pampari2018emrqa} to create a QA dataset from MADE 1.0 \citep{jagannatha2019overview}. This allows us to have two QA datasets with common question forms, but different text distributions. In this experimental setup we can mimic the evaluation of calibration methods in a real-world scenario, where the task specifications  may remain the same but the underlying text source changes. 
Details about dataset pre-processing and construction are provided in the Appendix.

The entity of interest for QA is the top-$k$ answer span predictions.  We use the ``lm'' perplexity as a feature in this experiment to analyze its behaviour in out-of-domain evaluations. We use a $2$ layer unidirectional LSTM to train a next word language model on the EMRQA passages. This language model is then used to compute the perplexity of a sentence for the ``lm'' input feature to the forecaster. We use the same baselines as the previous two tasks.

Based on Table \ref{table:QA-ECE}, our methods outperform the baselines by a large margin in both in-domain and out-of-domain experiments. Value of  \textit{heuristic-$k$}  is 3 for all +Rank+Var+topk forecasters. We show ablation and baseline results for $k=3$ only. However, no other forecasters for any $k\in \{2,3\}$ outperform our best forecasters in Table \ref{table:QA-ECE} . Our models are evaluated on SQuAD1.1 dev set, and test sets from EMRQA and MADE 1.0. They show consistent improvements in ECE and Exact Match Accuracy.
\begin{figure*}[htp]
    \centering
    \includegraphics[width=0.9\textwidth]{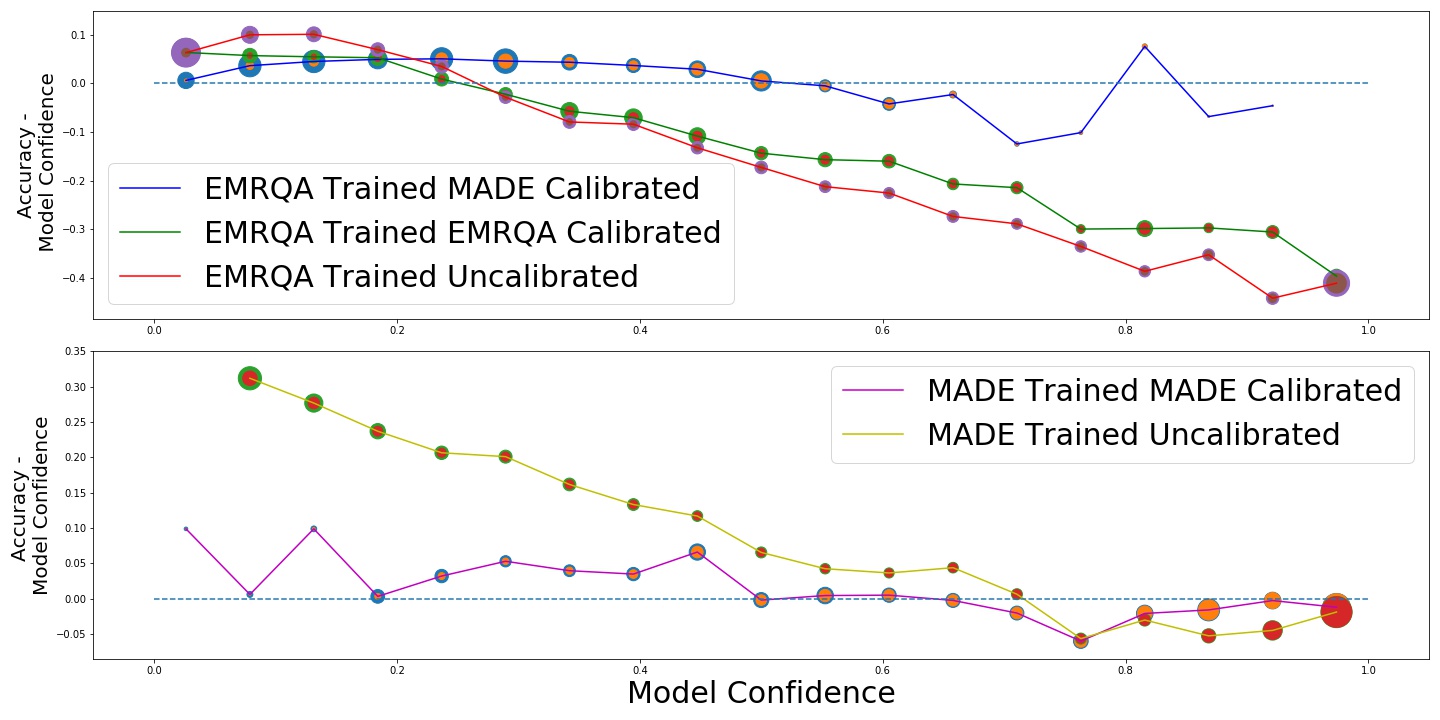}
    \caption{Modified reliability plots (\textit{Accuracy - Confidence} vs \textit{Confidence}) on MADE 1.0 QA test. The dotted horizontal line represents perfect calibration. Scatter point diameter denotes bin size. The inner diameter of the scatter point denotes the number of positive events in that bin.}
    \label{fig:oodcalibration}
\end{figure*}
\section{Discussion}
Our proposed methods outperform the baselines in most tasks and are almost as competitive in others. 
\paragraph{Features and top-$k$ samples:}  The inclusion of top-$k$ features improve the performance in almost all tasks when the rank of the prediction is included. We see large increases in calibration error when the top-$k$ prediction samples are included in forecaster training without including the rank information in tasks such as CoNLL NER and MADE 1.0 QA. This may be because the $k=1,2,3$ predictions may have similar model confidence and uncertainty values. Therefore a more discriminative signal such as rank is needed to prioritize them. For instance, the difference between probabilities of $k=1$ and $k=2$ MAP predictions for POS tagging may differ by only one or two tokens. In a sentence of length $10$ or more, this difference in probability when normalized by length would account to very small shifts in the overall model confidence score. Therefore an additional input of rank $k$ leads to a substantial gain in performance for all models in POS. 

Our task-agnostic scheme of ``Rank+Var+topk'' based forecasters consistently outperform or stay competitive to other forecasting methods. However, results from task-specific features such as ``lm'' and ``len'' show that use of task-specific features can further reduce the calibration error. Our domain shift experimental setup has  the same set of questions in  both in-domain and out-of-domain datasets. Only the data distribution for the answer passage is different. However, we do not observe an improvement in out-of-domain performance by using ``lm'' feature. A more detailed analysis of task-specific features in QA with both data and question shifts is required.  We leave further investigations of such schemes as our future work.

\begin{figure}[htp]
    \centering
    \includegraphics[width=0.5\textwidth]{{{conllexample}}}
    \caption{An example of named entity span from CoNLL dataset. Rank is $k^{th}$ rank from  top-$k$ MAP inference (Viterbi decoding). Mean Prob and Std is the mean and standard deviation of length-normalized probabilities (geometric mean of marginal probabilities for each token in the span). Calibrated confidence is the output of  \textit{Rank+Var+ln+top3}.
    }
    \label{fig:conllexample}
\end{figure}

\paragraph{Choice of k is important}: The optimal choice of $k$ seems to be strongly dependent on the inherent properties of the tasks and its output event set. In all our experiments, for a specific task all +Rank+Var+top$k$ forecasters exhibit consistent behaviours with respect to the choice of $k$. In POS experiments, \textit{heuristic-$k$} $=2$. In all other tasks, \textit{heuristic-$k$} $=3$. Our \textit{heuristic-$k$} models are the best performing models, suggesting that the heuristic described in Section \ref{sec:forecaster} may generalize to other tasks as well.

\paragraph{Re-scoring}: We show that using our forecaster confidence to re-rank the entities of interest leads to a  modest boost in model performance for the NER and QA tasks. In POS no appreciable gain or drop in performance was observed for $k=2$. We believe this may be due to the already high token level accuracy (above 97\%) on Penn Treebank data. Nevertheless, this suggests that our re-scoring does not lead to a degradation in model performance in cases where it is not effective.

Our forecaster re-scores the top-$k$ entity confidence scores based on model uncertainty score and entity-level features such as entity lengths. Intuitively, we want to prioritize predictions that have low uncertainty over high uncertainty predictions, if their uncalibrated confidence scores are similar. We provide an example of such re-ranking in Figure \ref{fig:conllexample}. It shows a named entity span predictions for the correct span ``Such''.  The model $p_\theta$ produces two entity predictions ``off-spinner Such'' and ``Such''. The un-calibrated confidence score of  ``off-spinner Such'' is higher than ``Such'', but the variance of its prediction is higher as well. Therefore the \textit{+Rank+Var+ln+top3} re-ranks the second (and correct) prediction higher. It is important to note here that the variance of ``off-spinner Such'' may be higher just because it involves two token predictions as compared to only one token prediction in ``Such''. This along with the ``ln'' feature in \textit{+Rank+Var+ln+top3} may mean that the forecaster is also using length information along with uncertainty to make this prediction.  However, we see similar improvements in QA tasks, where the ``ln'' feature is not used, and all entity predictions involve two predictions (span start and end index predictions). These results suggest that use of uncertainty features are useful in both calibration and re-ranking of predicted structured output entities.

\paragraph{Out-of-domain Performance}:
Our experiments testing the performance of calibrated QA systems on out-of-domain data suggest that our methods result in improved calibration on unseen data as well. Additionally, our methods also lead to an improvement in system accuracy on out-of-domain data, suggesting that the mapping learned by the forecaster model is not specific to a dataset. However, there is still a large gap between the calibration error for within domain and out-of-domain testing. This can be seen in the reliability plot shown in Figure \ref{fig:oodcalibration}. The number of samples in each bin are denoted by the radius of the scatter point. The calibrated models shown in the figure corresponds to ``+Rank+Var+lm+top3' forecaster calibrated using both in-domain and out-of-domain validation datasets for forecaster training. We see that out-of-domain forecasters are over-confident and this behaviour is not mitigated by using data-uncertainty aware features like ``lm''. This is likely due to a shift in model's prediction error when applied to a new dataset. Re-calibration of the forecaster using a validation set from the out-of-domain data seems to bridge the gap. However, we can see that the \textit{sharpness} \citep{kuleshov2015calibrated} of \textit{out-of-domain trained, in-domain calibrated} model is much lower than that of \textit{in-domain trained, in-domain calibrated} one. Additionally, a validation dataset is often not available in the real world. Mitigating the loss in calibration and sharpness induced by out-of-domain evaluation is an important avenue for future research.

\paragraph{Uncertainty Estimation}: We use MC-Dropout as a model (\textit{epistemic}) uncertainty estimation method in our experiments. However, our method is not specific to MC-Dropout, and is compatible with any method that can provide a predictive distribution over token level outputs. As a result any bayesian or ensemble based uncertainity estimation method \citep{welling2011bayesian,lakshminarayanan2017simple, ritter2018scalable} can be used with our scheme.  In this work, we do not investigate the use of \textit{aleatoric} uncertainty for calibration. Our use of language model features is aimed at accounting for \textit{distributional} uncertainty instead of \textit{aleatoric} uncertainty \citep{gal2016uncertainty,malinin2018predictive}. Investigating the use of different types of uncertainty for calibration remains as our future work.

\section{Conclusion}
We show a new calibration and confidence based re-scoring scheme for structured output entities in NLP. We show that our calibration methods outperform competitive baselines on several NLP tasks. Our task-agnostic methods can provide calibrated model outputs of specific entities instead of the entire label sequence prediction. We also show that our calibration method can provide improvements to the trained model's accuracy at no additional training or data cost. Our method is compatible with modern NLP architectures like \textsc{bert}. Lastly, we show that our calibration does not over-fit on in-domain data and is capable of generalizing the calibration to out-of-domain datasets.
\section*{Acknowledgement}
Research reported in this publication was supported by the National Heart, Lung, and Blood Institute (NHLBI) of the National Institutes of Health under Award Number R01HL125089.

\bibliography{acl2020}
\bibliographystyle{acl_natbib}

\clearpage
\appendix

\section{Appendices}
\label{sec:appendix}

\begin{algorithm*}

  \caption{ Forecaster construction for model $p_\theta$ with max rank $k_{max}$.\\
  \textbf{\textit{Input}}: Uncalibrated model $p_\theta$ , Validation Dataset  $\mathbb{D}= \{(x^{(i)},y^{(i)}\}_{i=0}^{|\mathbb{D}|}$ , $k_{max}$. \\
  \textbf{\textit{Output}}: Forecaster $F_y$}
  \label{alg:forecaster}

\Fn{Get-Forecaster ($p_\theta$, $\mathbb{D}$, $k_{max}$)}{
\SetAlgoLined
\For{$i\gets0$ \KwTo $|\mathbb{D}|$}{

$\mathcal{I}(x^{(i)}) \gets \text{Get-Candidate-Events}(p_\theta,x^{(i)},k_{max})$ \\
$\mathbb{D}_{train} \gets \{(x^{(i)},c,\textit{E}): c = \mathbbm{1}(y^{(i)} \in \textit{E})$ , $ \forall \textit{E} \in \mathcal{I}(x) \}$

$\mathcal{I}_{k=1}(x^{(i)}) \gets \text{Get-Candidate-Events}(p_\theta,x^{(i)},1)$\\
$\mathbb{D}_{val} \gets \{(x^{(i)},c,\textit{E}): c = \mathbbm{1}(y^{(i)} \in \textit{E})$ , $ \forall \textit{E} \in \mathcal{I}_{k=1}(x) \}$

}
Train Forecasters $F^{(k)}_y$ for $k=\{1,...,k_{max}\}$ using $\mathbb{D}_{train}$ \\
$F_y \gets F^{(k)}_y$ with minimum ECE on $\mathbb{D}_{val}$ \\
\Return $F_y$}

\Fn{Get-Candidate-Events ($p_\theta$,$x$,$k_{max}$)}{
    Construct top-$k_{max}$ label sequences using MC-Dropout average of $p_\theta(x)$ logits. \\
    Extract relevant entity set $\phi(x)$ from top-$k_{max}$ label sequences. \\
    $\mathcal{I}(x) \gets$ Events corresponding to entities in $\phi(x)$. \\
    \KwRet $\mathcal{I}(x)$\;
}
\end{algorithm*} 

\subsection{Algorithm Details:} 
The forecaster construction algorithm is provided in Algorithm \ref{alg:forecaster}. The candidate events in Algorithm \ref{alg:forecaster} are obtained by extracting top-$k$ label sequences for every output. The logits obtained from $p_\theta$ are averaged over 10 MC-Dropout samples before being fed into the final output layer. We use the validation dataset from the task's original split to train the forecaster. The validation dataset is used to construct both training and validation split for the forecaster. The training split contains all top-$k$ predicted entities. The validation split contains only top-$1$ predicted entities.

\subsection{Evaluation Details}
We use the expected calibration error (ECE) score defined by \citep{naeini2015obtaining} to evaluate our calibration methods. Expected calibration error is a score that estimates the expected absolute difference between model confidence and accuracy. This is calculated by binning the model outputs into $N$ ($N=20$ for our experiments) bins and then computing the expected calibration error across all bins. It is defined as \begin{equation}
    ECE= \sum_{i=0}^{N} \frac{|B_i|}{n}\lvert acc(B_i)-conf(B_i) \rvert , 
\end{equation} 
where $N$ is the number of bins, $n$ is the total number of data samples, $B_i$ is the $i^{th}$ bin. The functions $acc(.)$ and $conf(.)$ calculate the accuracy and model confidence for a bin.
\subsection{Implementation Details}
We use AllenNLP's wrapper with HuggingFace's Transformers code \footnote{https://github.com/huggingface/transformers} for our implementation\footnote{The code for forecaster construction is available at https://github.com/abhyudaynj/ StructuredPredictionCalibrationNLP}. We use \textit{\textsc{bert}-base-cased} \citep{wolf2019transformers} weights as the initialization for general-domain datasets and \textit{bio-\textsc{bert}} weights \citep{lee2019biobert} as the initialization for clinical datasets. We use cased models for our analysis, since bio-\textsc{bert}\citep{lee2019biobert} uses cased models. A common learning rate of 2e-5 was used for all experiments. We used validation data splits provided by the datasets. In cases where the validation dataset was not provided, such as MADE 1.0, EMRQA or SQuAD1.1, we use 10\% of the training data as the validation data. We use a patience of 5 for early stopping the model, with each epoch consisting of 20,000 steps. We use the final evaluation metric instead of negative log likelihood (NLL) to monitor and early stop the training. This is to reduce the mis-calibration of the underlying $p_\theta$ model, since \citet{guo2017calibration} observe that neural nets overfit on NLL. The implementation for each experiment is provided in the following subsections. 

\subsubsection{Part-of-speech experiments}
We evaluate our method on the Penn Treebank dataset \cite{marcus1994penn}. Our experiment uses the standard training (1-18), validation(19-21) and test (22-24) splits from the WSJ portion of the Penn Treebank dataset. The un-calibrated output of our model for a candidate label sequence is estimated as 
\begin{equation}
    \hat{p}=\frac{1}{M}\sum_{MC-Dropout}p_\theta(y_1,y_2, ... y_L|x)^{\frac{1}{L}},
\end{equation} where $M$ is the number of dropout samples. The $L^{th}$ root accounts for different sentence lengths. Here $L$ is the length of the sentence. We observe that this kind of normalization improves the calibration of both baselines and proposed models. We do not normalize the probabilities while reporting the ECE of uncalibrated models. We use two choice of $p_\theta$ models, namely \textsc{bert} and \textsc{bert}+CRF. \textsc{bert}  only model adds a linear layer to the output of \textsc{bert} network and uses a softmax activation function to produce marginal label probabilities for each token. \textsc{bert}+CRF uses a CRF layer on top of unary potentials obtained from the \textsc{bert} network outputs. 

We use Platt Scaling \citep{platt1999probabilistic} as the baseline calibration model. Our Platt scaling model uses the MC-Dropout average of length normalized probability output of the model $p_\theta$ as input. The lower variance and length invariance of this input feature make Platt Scaling a very strong baseline. We also use a ``Calibrated Mean'' baseline using Gradient Boosted Decision Trees as our estimator with the same input feature as Platt.

\subsubsection{NER Experiments}
For CoNLL dataset, ``testa'' file was reserved for validation data and ``testb'' was reserved for test. For MADE 1.0 \citep{jagannatha2019overview}, since validation data split was not provided we randomly selected 10\% of training data as validation data. The length normalized marginal probability for a span starting at $i$ and of length $l$ is estimated as 
\begin{equation}
    \hat{p}=\frac{1}{M}\sum_{MC-Dropout}p_\theta(y_i,y_2, ... y_{i+l-1}|x)^{\frac{1}{l}}.
\end{equation}

We use this as the input to both the baseline and proposed models. We observe that this kind of normalization improves the calibration of baseline and proposed models. We do not normalize the probabilities while reporting the ECE of uncalibrated models. We use BIO-tags for training. While decoding, we also allow spans that start with ``I-'' tag.

\subsubsection{QA experiments}
We use three datasets for our QA experiments, SQAUD 1.1, EMRQA and MADE 1.0. Our main aim in these experiments is to understand the behaviour of calibration and not the complexity of the tasks themselves. Therefore, we restrict the passage lengths of EMRQA and MADE 1.0 datasets to be similar to SQuAD1.1. We pre-process the passages from EMRQA to remove unannotated answer span instances and reduce the passage length to 20 sentences. EMRQA provides multiple question templates for the same question type (referred to as logical form in \citet{pampari2018emrqa}). For each annotation, we randomly sample 3 question templates for our QA experiments. This is done to ensure that question types that have multiple question templates are not over-represented in the data. For example, the question type for ``'Does he take anything for her |problem|'' has 49 available answer templates, whereas ``How often does the patient take |medication|'' only has one. So for each annotation, we sample 3 question templates for a question type. If the question type does not have 3 available templates, we up-sample. For more details please refer to \citet{pampari2018emrqa}.

EMRQA is a QA dataset constructed from named entity and relation annotations from clinical i2b2 datasets consisting of adverse event, medication and risk related questions \cite{pampari2018emrqa}. We aim to also test the performance of our calibration method on out-of-domain test data. To do so, we construct a QA dataset from the clinical named entity and relation dataset MADE 1.0, using the questions and the dataset construction procedure followed in EMRQA. This allows us to have two QA datasets with common question forms, but different text distributions. This experimental setup enables us to evaluate how a QA system would perform when deployed on a new text corpus. This corresponds to the application scenario where a fixed set of questions (such as Adverse event questionnaire \citep{naranjo1981method}) are to be answered for clinical records from different sources. Both EMRQA and MADE 1.0 are constructed from clinical documents. However, the documents themselves have different structure and language due to their different clinical sources, thereby mimicking the real-world application scenarios of clinical QA systems. 

\paragraph{MADE QA Construction} MADE 1.0 \citep{jagannatha2019overview} is an NER and relation dataset that has similar annotation to ``relations'' and ``medication'' i2b2 datasets used in EMRQA. EMRQA uses an automated procedure to construct questions and answers from NER and relation annotations. We replicate the automated QA construction followed by \citet{pampari2018emrqa} on MADE 1.0 dataset to obtain a corresponding QA dataset for the same. For this construction, we use question templates that use annotations that are common in both MADE 1.0 and EMRQA datasets. Examples of common questions are in Table \ref{table:MADEQ}. A full list of questions in MADE 1.0 QA is in ``question\_templates.csv'' file included in supplementary materials. The dataset splits for EMRQA and MADE QA are provided in Table \ref{table:MADE_EMRQA}. 

\begin{table*}[ht]
\centering
\begin{tabular}{p{0.15\linewidth}p{0.15\linewidth}p{0.7\linewidth}}

\textbf{Input} & \textbf{Output} & \textbf{Example Question Form}\\
\hline \hline 
Problem & Treatment & How does the patient manage her |problem|\\
Treatment & Problem & Why is the patient on |treatment|\\
Problem & Problem & Has the patient ever been diagnosed or treated for |problem|\\
Drug & Drug & Has patient ever been prescribed |medication|\\
\hline
\end{tabular}
\caption{\label{table:MADEQ} Examples of questions that are common in EMRQA and MADE QA datasets.}
\end{table*}

\begin{table*}[ht]
\centering
\begin{tabular}{p{0.4\linewidth}p{0.2\linewidth}p{0.2\linewidth}p{0.2\linewidth}}

\textbf{Dataset Name} & \textbf{Train} & \textbf{Validation} & \textbf{Test}\\
\hline \hline 
EMRQA & 74414 & 8870& 9198\\
MADE QA & 99496&14066&21309\\
\hline
\end{tabular}
\caption{\label{table:MADE_EMRQA} Dataset size for the MADE dataset QA pairs that were constructed using guidelines from EMRQA. EMRQA dataset splits are also provided for comparison.}
\end{table*}

\paragraph{Forecaster features} Since we only consider single-span answer predictions, we require a constant number of predictions ( answer start and answer end token index), for this task. Therefore we do not use the ``ln'' feature in this task.  The uncalibrated probability of an event is normalized as follows and then used as input to all calibration models.
\begin{equation}
\hat{p}=\frac{1}{M}\sum_{MC-Dropout}p_\theta(y_{start},y_{end}|x)^{1/2}   
\end{equation}
Unlike the previous tasks, extractive QA with single-span output does not have a varying number of output predictions for each data sample. It always only predicts the start and end spans. Therefore using length normalized (where length is always 2) uncalibrated output does not significantly affect the calibration of baseline models. However, we use the length-normalized uncalibrated probability as our input feature to keep our base set of features consistent throughout the tasks. Additionally, in extractive QA tasks with non-contiguous spans, the number of output predictions can vary and be higher than 2. In such cases, based on our results on POS and NER, the length-normalized probability may prove to be more useful. The ``Var'' feature and ``Rank'' feature is estimated as described in previous tasks.


\end{document}